\documentclass[10pt,twocolumn,letterpaper]{article}

\usepackage[pagenumbers]{cvpr}

\usepackage{graphicx}
\usepackage{amsmath}
\usepackage{amssymb}
\usepackage{booktabs}
\usepackage{float}
\usepackage{pifont}

\usepackage{xcolor}
\usepackage{stfloats}

\usepackage[pagebackref,breaklinks,colorlinks]{hyperref}
\usepackage{graphicx}
\usepackage{lipsum}

\usepackage[capitalize]{cleveref}
\crefname{section}{Sec.}{Secs.}
\Crefname{section}{Section}{Sections}
\Crefname{table}{Table}{Tables}
\crefname{table}{Tab.}{Tabs.}

\begin{document}

\title{PointSt3R: Point Tracking through 3D Grounded Correspondence}

\author{Rhodri Guerrier$^{1}$ \quad Adam W. Harley$^{2}$ \quad Dima Damen$^{1}$ \\
$^{1}$University of Bristol \quad
$^{2}$Meta Reality Labs Research\\
\small{\url{http://rhodriguerrier.github.io/PointSt3R}}
}
\maketitle

\begin{figure*}[!th]
    \centering
   \includegraphics[width=1.0\linewidth]{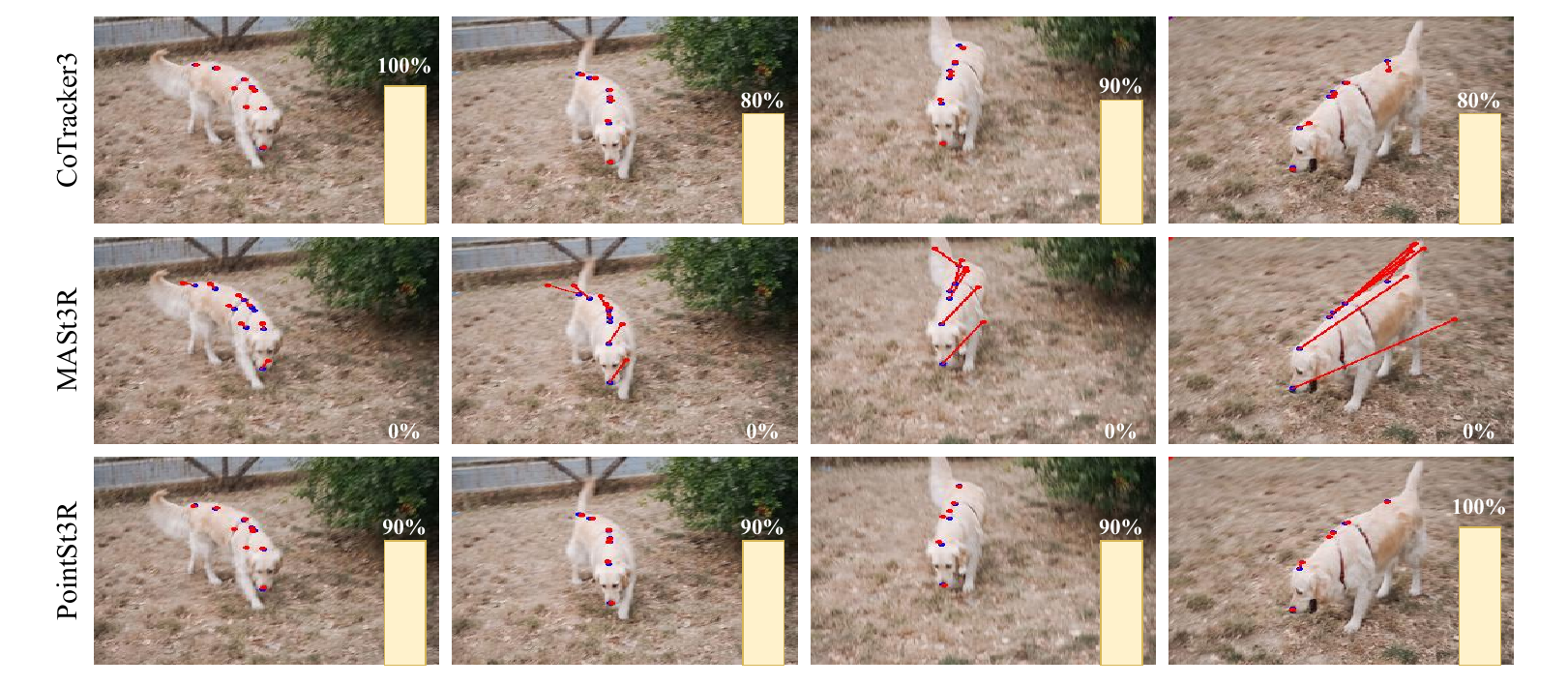}
   \vspace*{-2em}
\caption{\textbf{Comparison of approaches}: CoTracker3~\cite{karaev2024cotracker3}, MASt3R~\cite{leroy2024grounding} and PointSt3R, on four selected timesteps of a TAP-Vid-DAVIS~\cite{doersch2022tap} video. The blue dots represent the ground truth positions, whilst the red represent the predictions of the models. Where MASt3R clearly fails to track the dynamic scene, PointSt3R recovers accuracy, performing on par with CoTracker3. The yellow bars represent the visibility accuracy per frame. MASt3R has $0\%$ as it has no ability to predict visibility, whilst PointSt3R produces comparable results to CoTracker3.}
    \vspace*{-14pt}
  \label{fig:main_qual}
\end{figure*}

\begin{abstract}
  Recent advances in foundational 3D reconstruction models, such as DUSt3R and MASt3R, have shown great potential in 2D and 3D correspondence in static scenes. In this paper, we 
  propose to adapt them for the task of point tracking through 3D grounded correspondence. 
  We first demonstrate that these models are competitive point trackers when focusing on static points, present in current point tracking benchmarks ($+33.5\%$ on EgoPoints vs. CoTracker2).
  We propose to combine the reconstruction loss with training for dynamic correspondence along with a visibility head, and fine-tuning MASt3R for point tracking using a relatively small amount of synthetic data. 
  Importantly, we only train and evaluate on pairs of frames where one contains the query point, effectively removing any temporal context.
  
  Using a mix of dynamic and static point correspondences, we achieve competitive or superior point tracking results on four datasets
  (e.g. competitive on TAP-Vid-DAVIS
  73.8 $\delta_{avg}$ / 85.8\% occlusion acc. for PointSt3R compared to 75.7 / 88.3\% for CoTracker2; and significantly outperform CoTracker3 on EgoPoints 61.3 vs 54.2 and RGB-S 87.0 vs 82.8).
  We also present results on 3D point tracking along with several ablations on training datasets and percentage of dynamic correspondences.
  \label{fig:intro}
\end{abstract}

\vspace{-8pt}
\section{Introduction}
\label{sec:intro}

The task of point tracking, or pixel-level correspondence along frames of a video, has attracted renewed interest over the last few years. This is due in part to successes in supervised training that emphasise the use of temporal information \cite{zheng2023pointodyssey, karaev2024cotracker, karaev2024cotracker3}. Models have utilised temporal convolutions and attention operations, %
with synthetic training datasets, to produce impressive long-term tracking accuracy. Alternative methods based on multi-view correspondence~\cite{sun2021loftr} and optical flow~\cite{teed2020raft-172} have lagged behind, particularly in long dynamic scenes with occlusions. %

The recent wave of DUSt3R-inspired~\cite{wang2024dust3r} models focused on 3D reconstruction from unposed pairs of images~\cite{wang2024dust3r, leroy2024grounding} has suggested an interesting new approach for general-purpose point correspondence. These models are able to compute 3D locations for each pixel in a pair of images, making them potentially viable for dense point correspondence at arbitrary time differences. Furthermore, it has been shown that they can be easily modified with new regression heads to compute a number of related downstream tasks~\cite{leroy2024grounding, feng2025st4rtrack, jin2024stereo4d, sucar2025dynamic, wang2025vggt}. 
Specifically, MASt3R~\cite{leroy2024grounding}, which extends DUSt3R's~\cite{wang2024dust3r} static 3D reconstruction to feature-based pixel correspondence estimation, has shown that learning 3D geometry and feature matching together benefits both tasks, and can compete well with models designed solely for multi-view correspondence~\cite{sun2021loftr, sarlin2020superglue, lindenberger2023lightglue}.

Although DUSt3R and MASt3R show very impressive reconstruction and correspondence performance on static scenes, they fail to generalise to the more extreme task of corresponding points in dynamic scenes. This has led to a whole host of works~\cite{sucar2025dynamic, feng2025st4rtrack, chen2025easi3r, zhang2024monst3r} dedicated to introducing additional components to solve the problems posed by dynamic scenes, such as static/dynamic segmentation, 3D tracking, and time-aware reconstruction. Interestingly, MonSt3R~\cite{zhang2024monst3r} has shown that estimating 3D pointmaps in dynamic scenes is not necessarily an intrinsic failure of DUSt3R~\cite{wang2024dust3r} itself, but rather a symptom of the training regime. By fine-training DUSt3R with data that includes moving and deformable objects, they show that dynamic reconstruction can be greatly improved without any change to the architecture of the underlying model. In this work, we take MASt3R as a starting point, due to its feature matching abilities, and examine what dynamic scene training provides to its feature-based point correspondence.
This allows us to formulate point tracking as a 3D grounded correspondence task, which is attractive as it does not require temporal context or depth whilst simultaneously being able to track in both 2D and 3D.

Our contributions are as follows.\\ (1) We evaluate the performance of MASt3R~\cite{leroy2024grounding} to effectively correspond static and dynamic pixels using point tracking benchmarks.\\ (2)~We propose a new fine-tuning strategy that utilises annotated pairs of images taken at large temporal strides from existing 3D point tracking datasets to produce PointSt3R (\underline{Point} tracking by \underline{St}ereo \underline{3}D \underline{R}econstruction), a model designed for both 3D reconstruction and correspondence of dynamic scene points, which dramatically outperforms the original MASt3R~\cite{leroy2024grounding} baseline on point tracking. \\ (3) We show that PointSt3R is able to achieve competitive point tracking performance, in both 2D and 3D, without any knowledge of temporal context and depth during inference, effectively reformulating point tracking as a 3D grounded correspondence problem.

Fig~\ref{fig:main_qual} shows an example of dynamic point tracking using: (1) the competitive point tracking model \textbf{CoTracker3}~\cite{karaev2024cotracker3}, which heavily uses temporal information to track query points; (2)  \textbf{MASt3R}~\cite{leroy2024grounding}, which fails to track dynamic points; and (3) our proposed \textbf{PointSt3R} that recovers the performance of point trackers and visibility accuracy (yellow bars) using 3D grounded correspondence on pairs of frames across a video, \textit{without any temporal modelling}.

\section{Related Work}
\label{sec:related}

\noindent \textbf{Point Tracking.} Unlike classic optical flow models, which try to calculate instantaneous pixel velocities, point tracking attempts to calculate long-term trajectories~\cite{sand2008particle}. This task was recently re-popularised by the work of PIPs~\cite{harley2022particle}, and the TAP-ViD benchmark~\cite{doersch2022tap}. Focusing on tracking independent points, PIPs was one of the first to utilise synthetic data and a deep learning model that calculates correlation maps across a multi-frame inference window in order to produce tracks through occlusions. Soon after PIPs, the TAP-ViD datasets~\cite{doersch2022tap} provided a number of benchmarks for evaluating models on diverse real-world videos. Point tracking performance was improved with introduction of attention between points, and training with a sliding window, as introduced in CoTracker~\cite{karaev2024cotracker}. A number of new models have since appeared with different methods such as 4D correlations \cite{cho2024local}, optimisation-based techniques \cite{tumanyan2024dino, wang2023tracking}, transformer-based models \cite{li2024taptr, li2024taptrv2}, wider spatial~\cite{bian2023context} or temporal~\cite{zheng2023pointodyssey} context  and bootstrapping on real-world data~\cite{sun2024refining, doersch2024bootstap, karaev2024cotracker3}. More recently, point tracking has moved to 3D. Works include SpaTracker~\cite{xiao2024spatialtracker}, which utilises pre-trained depth estimators and triplane representations for efficiency, and DELTA~\cite{ngo2024delta} which focuses on low resolution tracking, transformer-based global attention and upsampling techniques.  %

Different from these works, our work takes a simpler approach to point tracking: we take pairs of frames, encode them jointly, estimate their relative camera and represent all points in one image in the world coordinate frame of the second, as done in standard 3D reconstruction. 
We then use nearest-neighbour correspondence from the feature-based pixel correspondence head as the base of point tracking.

\vspace{1.0em}\noindent \textbf{3D Reconstruction \& Tracking.} Methods for Structure from Motion \cite{agarwal2011building} and Simultaneous Localisation and Mapping \cite{davison2007monoslam} have traditionally relied on optical flow or point tracks being computed with an intermediate module~\cite{ozyecsil2017survey,wang2024vggsfm, li2025megasam}, but some recent methods have tackled correspondence and reconstruction jointly.
DUSt3R~\cite{wang2024dust3r} introduced a powerful approach that produces a pair of 3D pointmaps for a pair of images, which accommodates a number of downstream tasks including 3D reconstruction, camera pose estimation, and depth estimation. As DUSt3R is limited to static scenes, several follow-up works have since followed focusing on dynamic environments and extending the temporal context of the model \cite{zhang2024monst3r, wang2025continuous, wang20243d}. MASt3R (Matching And Stereo
3D Reconstruction)~\cite{leroy2024grounding}
showed that correspondence could be achieved relatively easily with these powerful 3D reconstruction models by adding a feature head and re-training, which inspires our own approach.

Follow-up methods achieve correspondence by special re-designs of the architecture: St4RTrack~\cite{feng2025st4rtrack} and DPM~\cite{sucar2025dynamic} add new components to DUSt3R to enable timestamp and viewpoint to be disentangled.
VGGT~\cite{wang2025vggt} proposes a DUSt3R-inspired architecture to simultaneously regress camera poses and pointmaps, and employs an expensive multi-stage training regime to capture 3D scenes from multi-frame input.
VGGT also showed that the features from this backbone can be used as input to point tracking methods (particularly CoTracker).
Different from our work, their approach still uses a special-purpose point tracking architecture, while we simply use the features from our backbone and track points with nearest neighbour.
 
Our approach returns to the simplicity of MASt3R and attempts to add dynamics by introducing a dynamic matching loss and fine-tuning on dynamic point correspondences from synthetic datasets. Our work is the first to use correspondence solely for point tracking.
We introduce our approach next.

\section{Our Method: PointSt3R}

In our approach, we formulate point tracking as a simple 2-frame correspondence task. We build on the foundational model MASt3R that achieves state-of-the-art performance on 2-view static point correspondence.
We propose to make MASt3R into a general-purpose 2-frame correspondence model by fine-tuning it with a point matching loss in dynamic scenes.
To highlight our connection to DUSt3R and MASt3R, we name our approach \underline{Point} tracking by \underline{St}ereo \underline{3}D \underline{R}econstruction (PointSt3R). In this section we describe our approach in detail, and demonstrate how it can be used for both 2D and 3D point tracking.

\subsection{Point Tracking as a Correspondence Task}
\label{sec:problem_definition}
In its simplest form, point tracking expects two frames as input. 
The first is the frame with one or more query points (as pixel coordinates), and the second is the target frame where point correspondences should be estimated.
Note that point tracking aims to track any point, not only keypoints or easily identifiable points.
Dense (per-pixel) correspondence between the frames subsumes the task of tracking sparse queries.

To form dense correspondences between some query frame $I^0$ and some target frame $I^t$, we simply take each feature in the query frame, compute its cosine similarity to each feature in the target frame, and take the coordinate of the maximum. 
This provides us with predicted correspondences $\mathcal{M}_{t} = {(i,j)}$, where $i, j \in \mathbb{R}^2$ are 2D coordinates in the image plane that are predicted as corresponding. To estimate a complete trajectory for a query point, we simply repeat this operation across every possible target frame.
Note that the correspondences  need only be computed at the frame level, and all queries within a frame are tracked simultaneously.

While it is possible to do post processing on these estimates to remove outliers or smooth out the resulting trajectories, our goal is to test the limit of pairwise correspondence, forming a point tracker \textit{with no bells and whistles}.

\subsection{MASt3R Background}

\vspace{1pt}\noindent \textbf{Architecture.} MASt3R~\cite{leroy2024grounding} is a variant of the original DUSt3R~\cite{wang2024dust3r}. These methods take a pair of images $I^1$ and $I^2$ as input, and encode these individually into feature maps $H^1, H^2$ (with a siamese encoder), and then decode them jointly into feature maps $H'^1, H'^2$. These features are concatenated channel-wise as $[H^1, H'^1]$ and $[H^2, H'^2]$ to serve as inputs to the model heads. The heads output pointmaps $X^{1,1}$ and $X^{2,1}$, as well as confidence maps $C^1$, and $C^2$, where $X^{i,j} \in \mathbb{R}^{H\times W\times 3}$ is a 3D pointmap corresponding to the $H \times W$ sized image $I^i$ in the coordinate frame of image $I^j$, and $C^i$ is the confidence map associated with the pointmap.
Importantly, MASt3R extends DUSt3R~\cite{wang2024dust3r} by adding a head $\mathcal{H}_{\textrm{desc}}$ that produces descriptors $D^1, D^2 \in \mathbb{R}^{H \times W \times d}$ for cross-image matching:
\begin{equation}
D^1 = \mathcal{H}_{\textrm{desc}}([H^1,H^{'1}]),
\end{equation}
\begin{equation}
D^2 = \mathcal{H}_{\textrm{desc}}([H^2,H^{'2}]),
\end{equation}
where $[H^i,H'^i]$ are the concatenated encoder and decoder features, as done for the other heads.

\vspace{1.0em}\noindent \textbf{Training.} MASt3R~\cite{leroy2024grounding} is trained in a supervised manner on exclusively static scenes using two losses. The first is a confidence-aware regression loss on the 3D positions of pixels in both images. The (unweighted) regression loss is given by 
\begin{equation}
\ell_{\textrm{regr}} = || \frac{1}{z} X_i^{v,1} - \frac{1}{\hat{z}} \hat{X}_i^{v,1} ||,
\end{equation}
where $v \in \{1,2\}$ is the view and $i$ is a pixel index, and $z, \hat{z}$ are normalizing factors defined as mean distance of points from the origin. The confidence weighted regression is given by
\begin{equation}
\mathcal{L}_{\textrm{conf}} = \sum_{v \in \{1,2\}} \sum_{i \in V^{v}} C_{i}^{v}\ell_{\textrm{regr}}(v,i) - \alpha \log C_{i}^{v},
\end{equation}
which down-weights the regression loss on estimates with low confidence, while penalising low confidence overall.

The second loss is for training the correspondence matching. This is achieved with an infoNCE loss~\cite{oord2018representation},  applied for the ground-truth correspondences ${\hat{\mathcal{M}} = \{(i,j) | \hat{X}_{i}^{1,1} = \hat{X}_{j}^{2,1} \}}$:
\begin{align}
\mathcal{L}_{\textrm{match}}^{\textrm{static}} =
& - \sum_{(i,j) \in \hat{\mathcal{M}}} 
\bigg[
\log{\frac{s_{\tau}(i,j)}{\sum_{k \in \mathcal{P}^1} s_{\tau}(k,j)}} 
\nonumber \\[4pt]
& \qquad
+ \log{\frac{s_{\tau}(i,j)}{\sum_{k \in \mathcal{P}^2} s_{\tau}(i,k)}} 
\bigg],
\label{eq:loss_matching}
\end{align}

\begin{equation}
\textrm{with} \ s_{\tau}(i,j) = \exp[-\tau D^{1T}_{i} D^{2}_{j}],
\end{equation}
where $\mathcal{P}^1 = \{i | (i,j) \in \hat{\mathcal{M}} \}$ and $\mathcal{P}^2 = \{j | (i,j) \in \hat{\mathcal{M}} \}$ denote the subset of pixels that match across images (according to ground truth), and $\tau$ is temperature hyper-parameter. In practice we use a confidence-weighted version of this loss, sharing confidence maps with the regression loss. %

\begin{table}[]
\centering
\small
\resizebox{\linewidth}{!}{
\begin{tabular}{l|cc|cc}
\toprule
& \multicolumn{2}{c|}{Static} & \multicolumn{2}{c}{Dynamic} \\
\midrule
Model       & PointOdyssey & EgoPoints & PointOdyssey & EgoPoints \\
\midrule
PIPs++~\cite{zheng2023pointodyssey} & 34.5 & 47.6 & 22.7 & 20.0 \\
CoTracker2~\cite{karaev2024cotracker} & 42.1 & 45.2 & \underline{26.2} & \underline{20.2}  \\
CoTracker3~\cite{karaev2024cotracker3} & \textbf{82.6} & \underline{65.0} & \textbf{50.1} & \textbf{35.8} \\
MASt3R~\cite{leroy2024grounding}   & \underline{67.0} & \textbf{78.7} & 24.5 & 12.3  \\
\bottomrule
\end{tabular}}
\caption{Comparison of CoTracker2~\cite{karaev2024cotracker} and CoTracker3~\cite{karaev2024cotracker3} and MASt3R~\cite{leroy2024grounding} on PointOdyssey~\cite{zheng2023pointodyssey} and the EgoPoints~\cite{darkhalil2025egopoints} benchmarks, reported on labelled static and dynamic point subsets.}
\vspace*{-12pt}
\label{tab:baseline_static_v_dyn}
\end{table}

\vspace{1.0em}\noindent \textbf{MASt3R as a Point Tracker.}
As noted earlier, the MASt3R~\cite{leroy2024grounding} model is trained on static correspondences. By design, this makes it a potentially competitive point tracker for query points on the static environment, using our reformulation of point tracking as a correspondence problem~Sec~\ref{sec:problem_definition}.
We confirm this assumption on two datasets where points are reliably labelled as \textit{static} or \textit{dynamic}, as we will later discuss in Sec~\ref{sec:experiments}. %

As can be seen in Table \ref{tab:baseline_static_v_dyn}, MASt3R is a strong point tracker when it comes to static points, due to its correspondence training. It outperforms all point trackers on the static points of EgoPoints~\cite{darkhalil2025egopoints}, beating the state-of-art CoTracker3~\cite{karaev2024cotracker3} by $+13.7\%$. 
On the static subset of PointOdyssey~\cite{zheng2023pointodyssey} test set, MASt3R~\cite{leroy2024grounding} beats CoTracker2~\cite{karaev2024cotracker} by $+24.9\%$, but is below CoTracker3.

\begin{figure}[t]
  \centering
   \includegraphics[width=\linewidth]{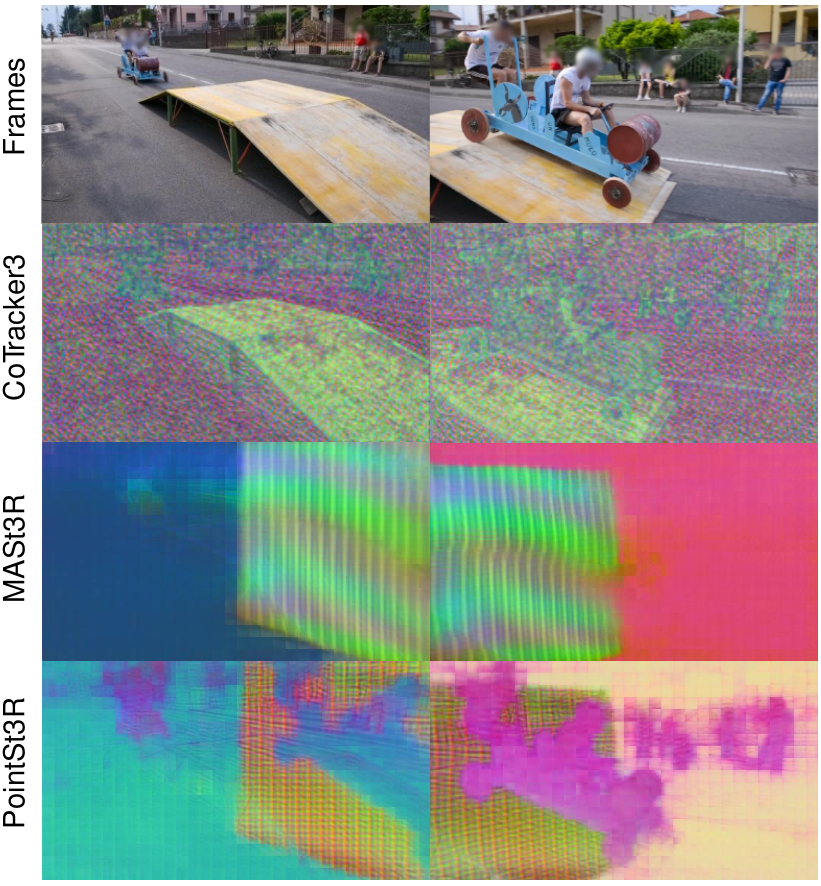}
   \vspace{-2em}
   \caption{PCA visualisation of feature maps extracted from CoTracker3~\cite{karaev2024cotracker3}, MASt3R~\cite{leroy2024grounding} and our PointSt3R on two frames of a TAP-Vid-DAVIS~\cite{doersch2022tap} video. CoTracker3 produces locally-sensitive features without global artifacts. MASt3R produces globally-sensitive features that are represent the 3D static background, with no unique features for dynamic objects. PointSt3R balances this by introducing local features and retaining 3D global features.}
   \vspace{-12pt}
   \label{fig:baseline_ft_comp}
\end{figure}

This ability to track static correspondences well is visualised in Fig.~\ref{fig:baseline_ft_comp}. Here we perform PCA on the two feature maps $D^1, D^2$ output by MASt3R for two overlapping frames of TAP-Vid-DAVIS~\cite{doersch2022tap}. It is clear that the static environment is tracked well as we can see the original grid of features mapped correctly to the left and warped for the change in pose. However, what is missing is unique features for dynamic objects, as the features of non-overlapping static areas are completely distinct in colour. For example, the cart is barely picked up in the first image, being confused with the features of the static region and is then lost in the second. This is in sharp contrast to the features extracted from CoTracker3~\cite{karaev2024cotracker3}. These features are more diverse and clearly capture the dynamic objects and static environments together.

Table \ref{tab:baseline_static_v_dyn} similarly confirms that, although a powerful static point tracker, MASt3R struggles greatly with dynamic correspondences. On PointOdyssey, it falls short of CoTracker3 by a large margin of $-25.6\%$ and similarly on EgoPoints by $-23.5\%$.

Based on these observations, we propose a new training strategy that converts MASt3R into a general-purpose correspondence model for both static and dynamic points. 
We show in Fig.~\ref{fig:baseline_ft_comp} (bottom) how we recover local features and dynamic objects (e.g. cart shows a corresponding purple colour).
We describe how PointSt3R is trained next.

\subsection{Training PointSt3R}
\vspace{-8pt}
\vspace{1.0em}\noindent \textbf{Architecture.}. Our hypothesis is that simply a new training regime can enable MASt3R to handle dynamic points.
This is inspired by prior work that fine-tuned DUSt3R for dynamic scene reconstruction (e.g., ~\cite{zhang2024monst3r}), but we note that these models do not have the ability to find pixel correspondences and are thus incapable of point tracking.

\vspace{1.0em}\noindent \textbf{Training.} We hypothesise that a slight change to the standard MASt3R training regime, to include dynamic points matching in training, should enable a boost in dynamic point tracking.
We thus introduce new loss terms for dynamic point matching: an infoNCE matching loss on dynamic points.
To model occlusion, we add a new head to each branch to predict the visibility of each pixel in the other image along with a cross-entropy visibility loss.

For a pixel correspondence between images from two timesteps, $i_{\textrm{img}},j_{\textrm{img}} \in \mathbb{R}^2$, we obtain their 3D camera-stabilized coordinates $i_{\textrm{world}}, j_{\textrm{world}} \in \mathbb{R}^3$.
If the point is static then $i_{\textrm{world}} = j_{\textrm{world}}$, and if it is dynamic then ${i_{\textrm{world}} \not= j_{\textrm{world}}}$. We can thus define static and dynamic matches as $\hat{\mathcal{M}}^{\textrm{static}} = \{(i,j) | i_{\textrm{world}} = j_{\textrm{world}} \}$ and $\hat{\mathcal{M}}^{\textrm{dynamic}} = \{(i,j) | i_{\textrm{world}} \not= j_{\textrm{world}} \}$.
We define $\mathcal{L}_{\textrm{match}}^{\textrm{dynamic}}$ to be equivalent to Eq.~\ref{eq:loss_matching}, but using $\hat{\mathcal{M}}^{\textrm{dynamic}}$ instead of $\hat{\mathcal{M}}^{\textrm{static}}$.
When training, we control the percentage of dynamic correspondences by simply adjusting ${r = \frac{\left| \hat{\mathcal{M}}^{\textrm{dynamic}} \right|}{\left| \hat{\mathcal{M}}^{\textrm{dynamic}} \right| + \left| \hat{\mathcal{M}}^{\textrm{static}} \right|}}$ .

We use the ground truth occlusion values for each point - i.e. if the point is visible in one frame but invisible (occluded or out of view) in the other, we predict the point as occluded in the visible head. We use a balanced cross-entropy loss for training the visibility heads, as the distribution is often skewed in the direction of visible points. We refer to this loss term as $\mathcal{L}_{CE}^{vis}$. The dynamic correspondence and visibility losses are summed with the original terms: 
\begin{equation}
\mathcal{L}_{\textrm{total}} = \mathcal{L}_{\textrm{conf}} + \alpha \mathcal{L}_{\textrm{match}}^{\textrm{static}} + \beta \mathcal{L}_{\textrm{match}}^{\textrm{dynamic}} + \gamma \mathcal{L}_{CE}^{vis},
\label{eq:final_loss}
\end{equation}
where $\alpha, \beta, \gamma$ are the hyperparameters of the losses.

Training this loss requires new data with known dynamic correspondences: pairs of frames from dynamic scenes, with 2D correspondences, visibility indicators, camera parameters (intrinsics and extrinsics) and depth for each pixel. For this reason, we use synthetic data, which is a common source of supervision in point tracking and 3D reconstruction \cite{zheng2023pointodyssey, karaev2024cotracker, wang2024dust3r, leroy2024grounding}. 
The exact datasets used are described in the section \ref{sec:datasets}. 
Importantly, the training image pairs should have large strides between their timestamps in order to learn long-term correspondences suitable for long term point tracking of dynamic objects. More details about strides can be found in sections~\ref{sec:impl_details}. We also ablate the stride lengths in~\ref{sec:ablations}.

\vspace{1.0em}\noindent \textbf{PointSt3R as a 2D Point Tracker.}
 In point tracking, we are given a query point $i \in \mathbb{R}^2$ and a sequence of $T$ video frames $I^0, I^1,...,I^T$, where $I^i \in \mathbb{R}^{H \times W \times 3}$, and asked to produce a set of trajectory coordinates for each query point. As PointSt3R is inherently pairwise, %
 we feed the model $I^q$ and $I^t$, where $q$ is the timestep of the query and $t$ is any other timestep, and we extract the respective feature maps $D^q, D^t$. We then use the point feature  $D^q(i)$ %
 to compute cosine similarity against all  features in $D^t$, and we simply take the pixel position of the most similar feature as the target point position.
 As we noted earlier, we do not perform any post processing on this trajectory, in order to test the limits of pairwise correspondence for point tracking. 

\vspace{1.0em}\noindent \textbf{PointSt3R as a 3D Point Tracker.} As MASt3R~\cite{leroy2024grounding} indexes both the pointmap and correspondence heads with the pixel values, it is relatively straightforward to extend this to 3D point tracking. This task is defined as the estimation of a 3D trajectory in the coordinate system of the first frame's camera~\cite{koppula2024tapvid}. Specifically, we perform tracking with the same nearest-neighbour matching technique as for 2D, but now we return the corresponding 3D coordinates retrieved from our pointmaps.
Just as in our 2D point tracking formulation, we calculate the 3D trajectory by pairwise querying ($I^q$, $I^t$), and do not perform any post processing. 
An alternative way is to retrieve 2D correspondences first, then use the pixel depth from off-the-shelf depth estimators and the camera intrinsics matrix (or an estimate) to lift these to 3D. We compare both approaches.
\vspace{-8pt}
\section{Experiments}
\label{sec:experiments}

\subsection{Datasets}\label{sec:datasets} To  train PointSt3R we use a mixture of three datasets. First, we use PointOdyssey~\cite{zheng2023pointodyssey}, a large-scale synthetic dataset with deformable characters and real-world camera motion. Second, we use the CoTracker3~\cite{karaev2024cotracker3} version of the Kubric dataset~\cite{greff2022kubric}, a synthetic dataset mainly comprised of objects colliding with each other. Third, we use \mbox{DynamicReplica~\cite{karaev2023dynamicstereo}}, a depth estimation training and evaluation dataset that includes point tracks. As all three datasets include both 2D image point tracks and their corresponding 3D trajectories, we find dynamic points as those correspondences in 2D tracks that do not have the same 3D world coordinates across any pair of images. During training, we follow MASt3R by attempting to sample 4096 positive correspondences and 4096 negative samples for each sample. However, as the stride length is large, the number of correspondences often drops below 4096, so we pad with extra negatives.

\begin{table*}[]
\centering
\resizebox{0.8\linewidth}{!}{
\begin{tabular}{l|cccc|ccc}
\toprule
           &                \multicolumn{4}{c|}{All Points} & \multicolumn{3}{c}{Dynamic Points} \\
\cmidrule{2-8}
Model      & TAP-Vid-DAVIS & RoboTAP  & RGB-S & EgoPoints  & RoboTAP & RGB-S & EgoPoints   \\
\midrule
PIPs++~\cite{zheng2023pointodyssey} & 69.1  & 63.0 & 77.8 & 36.9  & 62.7 & 51.3 & 20.0   \\
CoTracker2~\cite{karaev2024cotracker} & 75.7 & 70.6 & \textbf{83.3}  & 35.5  & 66.9 & 53.7 & 20.2   \\
CoTracker3~\cite{karaev2024cotracker3} & \textbf{76.7}  & \textbf{78.8} & 82.8 & \textbf{54.2}  & \textbf{74.6} & \textbf{65.2}  & \textbf{35.8} \\
\hline
LoFTR~\cite{sun2021loftr}      & 22.8 & 24.8 & 28.8 & 9.2 & 11.4 & 18.6  & 4.8 \\
DINOv2~\cite{oquab2023dinov2}  & 13.4 & 15.1 & 15.9  & 5.8 & 10.7 & 11.0 & 4.8  \\
{MASt3R}~\cite{leroy2024grounding} & 38.5 & 71.6 & 73.2 & 53.5 & 42.5 & 39.4 & 12.3 \\
PointSt3R & \textbf{73.8} & \textbf{78.6} & \textbf{87.0} & \textbf{61.3} & \textbf{69.4} & \textbf{61.6} & \textbf{31.4} \\

\bottomrule
\end{tabular}
}
\caption{Tracking accuracy $\delta_{\text{avg}}\uparrow$ for PointSt3R, MASt3R, compared to popular point tracking and image correspondence methods on TAP-Vid-DAVIS~\cite{doersch2022tap}, RoboTAP~\cite{vecerik2024robotap}, RGB-S~\cite{doersch2022tap} and EgoPoints~\cite{darkhalil2025egopoints} using the ``first" query mode described in \cite{doersch2022tap}. 
We also report on dynamic points only for three datasets (see text for details).
We separate the temporally-informed trackers from the generic correspondence models (which operate pairwise) with a horizontal bar.
}
\label{tab:point_tracking_2d}
\vspace{-1.0em}
\end{table*}

For evaluation datasets, we report on five established point tracking benchmarks. (1) TAP-Vid-DAVIS~\cite{doersch2022tap}, a dataset of 30 real-world videos taken from YouTube and hand-annotated on mostly dynamic objects. (2) RoboTAP~\cite{vecerik2024robotap}, a dataset of roughly 250 real-world videos of robotic arms moving objects around a scene and (3) TAP-Vid-RGB-Stacking (RGB-S)~\cite{doersch2022tap} which is a synthetic version of RoboTAP with 50 videos. We also evaluate on the recently proposed (4)~EgoPoints~\cite{darkhalil2025egopoints}, a benchmark of 517 egocentric videos with an emphasis on large camera motion and point re-id. Finally, we include the mini validation dataset of (5)~PStudio from TAP-Vid-3D~\cite{koppula2024tapvid}, which is a studio-captured dataset with 3D trajectories on people and objects for 50 videos, to evaluate 3D point tracking.

\begin{table}[t]
\centering
\small
\begin{tabular}{ll}
\toprule
Model  & $OA$ \\ %
\midrule
CoTracker2~\cite{karaev2024cotracker} & 88.3 \\
CoTracker3~\cite{karaev2024cotracker3} & 90.2\\
PointSt3R                              & 85.8\\ %
\bottomrule
\end{tabular}
\caption{Occlusion accuracy ($OA$) on TAP-Vid-DAVIS}
\vspace*{-12pt}
\label{tab:point_tracking_vis}
\end{table}

As the MASt3R baseline struggles so much with dynamic points, we identify a subset of points within RoboTAP, RGB-S and PointOdyssey~\cite{zheng2023pointodyssey} that are certainly dynamic, and evaluate on these in addition to the evaluation on all points. Although RoboTAP \& RGB-S do not provide specific labels for dynamic points, a number of their videos have static cameras. Therefore, in order to produce the dynamic splits seen in Table \ref{tab:point_tracking_2d}, we sample points that move at least $10\%$ of the screen size on videos with no camera motion. This provides us with 906 and 515 tracks across 102 and 50 videos for RoboTAP and RGB-S dynamic splits, respectively. For EgoPoints, we use the available dynamic object labels and for PointOdyssey we randomly sample 50 static and 50 dynamic points per test video, and evaluate on 100 equally spaced frames from each video for efficiency.

The metric used exclusively for all 2D point tracking results is the average tracking accuracy $\delta_{\text{avg}}$. This is defined as in previous point tracking works \cite{zheng2023pointodyssey, karaev2024cotracker, karaev2024cotracker3} as the percentage of visible points tracked within $1, 2, 4, 8, 16$ pixels of the ground truth, averaged over the thresholds. We follow~\cite{doersch2022tap, karaev2024cotracker}, where point tracks are calculated for frames after the first appearance of each point.

\subsection{Implementation Details}\label{sec:impl_details} We freeze MASt3R's encoder and then fine-tune the decoders and heads with $10,000$ samples, evenly split between PointOdyssey~\cite{zheng2023pointodyssey}, Kubric~\cite{karaev2024cotracker3} and DynamicReplica~\cite{karaev2023dynamicstereo}. The weights are initialised using the pre-trained MASt3R weights. The additional visibility heads use the same MLP architecture as the feature heads, apart from the output dimension, which is set to 1, and are randomly initialised.
In Eq~\ref{eq:final_loss}, we set $\alpha, \beta$ to 0.075, as in~\cite{leroy2024grounding} code, and $\gamma$ to 1.0.
Following~\cite{zhang2024monst3r} temporal striding technique, where the sampling probability increases linearly with stride length, we use image pairs of temporal strides $[10,30,50,70,90,110,130,150,170]$ for PointOdyssey and DynamicReplica, and for Kubric (since the videos are shorter) we use strides $[10,20,...,90]$. A learning rate of $5\times 10^{-5}$ and a batch size of $16$ is used over 4 H100 GPUs. We set the ratio of dynamic points $r = 95\%$ and ablate this choice.  We follow the same training image resolutions and augmentations as in MASt3R~\cite{leroy2024grounding}. It takes roughly 12 hours to train PointSt3R for 50 epochs.
For all inference, we use the input resolution of $(512,384)$ and bilinear interpolation when indexing correspondence features.

\subsection{2D Point Tracking}
\label{sec:point_tracking_2d}

In Table \ref{tab:point_tracking_2d}, we compare PointSt3R and MASt3R, along with three recent point tracking methods, PIPs++~\cite{zheng2023pointodyssey}, CoTracker2~\cite{karaev2024cotracker} and CoTracker3~\cite{karaev2024cotracker3} (top). 
We also compare to popular image correspondence baselines, LoFTR~\cite{sun2021loftr} and DINOv2~\cite{oquab2023dinov2}, as well as MASt3R. 

\begin{figure}[t]
  \centering
   \includegraphics[width=\linewidth]{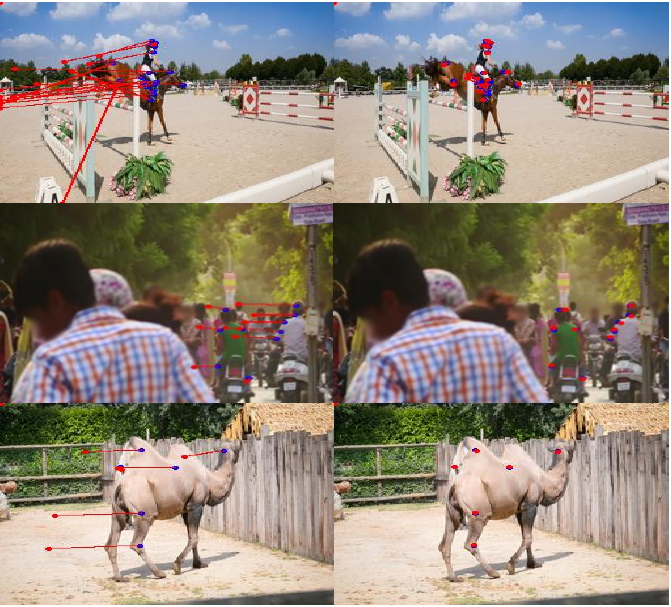}
   \vspace*{-1em}
   \caption{Comparison of MASt3R~\cite{leroy2024grounding} (left) and PointSt3R (right) tracks on TAP-Vid-DAVIS~\cite{doersch2022tap}. The blue dots represent the ground truth positions, whilst the red represent the predictions of the models. When prediction is far from GT, a red dash/line is visible. Larger errors are evident for MASt3R.}
   \vspace*{-12pt}
   \label{fig:qual_2}
\end{figure}

In TAP-Vid-DAVIS, although MASt3R~\cite{leroy2024grounding} is better than previous correspondence models LoFTR and DINOv2, it is much worse than point trackers, coming in behind PIPs++ by $-30.6\%$. However, our PointSt3R model is able to achieve competitive results when compared to point trackers. It boosts performance over the baseline by $+35.3\%$, outperforms PIPs++ by $+4.7\%$ and only falls short of state-of-the-art CoTracker2 and CoTracker3 by $-1.9\%$ and $-2.9\%$, respectively. 
We also report occlusion accuracy $OA$ in Table~\ref{tab:point_tracking_vis}. PointSt3R is able to achieve $85.8\%$ occlusion accuracy \textit{without temporal context} only slightly below CoTracker2 and CoTracker3 - $2.5\%$ and $4.4\%$, respectively.
Qualitative examples can be seen in Fig.~\ref{fig:main_qual} and~\ref{fig:qual_2}.

In RoboTAP, the MASt3R baseline does much better, due to the large amount of static points prevalent in these videos. PointSt3R is able to outperform PIPs++ and CoTracker2 by $+15.6\%$ and $+8.0\%$, respectively, and only falls short of CoTracker3 by $-0.2\%$. Although PointSt3R only outperforms the MASt3R baseline by $+7.0\%$, this gap is increased substantially on the dynamic split to $26.9\%$.

In TAP-Vid-RGB-Stacking (RGB-S), MASt3R similarly performs well due to the large amount of static points, as can be seen from the poor performance on the dynamic points. Training PointSt3R lifts performance by $+13.8\%$ and actually outperforms the best point tracker for this dataset, CoTracker2~\cite{karaev2024cotracker}, by $3.7\%$. In dynamic points, PointSt3R also outperforms CoTracker2 and is only $-3.6\%$ behind CoTracker3.

In EgoPoints, which focuses on challenging egocentric video with large camera motions and difficult re-ID situations, MASt3R actually outperforms all previous models when evaluating on all points, and PointSt3R improves this by $+7.8\%$. 
On dynamic points, the baseline MASt3R model underperforms the worst point tracker, PIPs++, by $-7.7\%$. PointSt3R is able to outperform CoTracker2 by $+11.2\%$, and is only $4.4\%$ below CoTracker3.

\begin{table}[t]
\centering
\small
\resizebox{0.8\linewidth}{!}{
\begin{tabular}{lcc}
\toprule
Model  & w G.T. & w/o G.T. \\ %
\midrule
CoTracker3~\cite{karaev2024cotracker3}+ZoeDepth & 80.3 & 66.2 \\
SpaTracker~\cite{xiao2024spatialtracker}+ZoeDepth & \textbf{81.6} & \textbf{67.1} \\
DELTA~\cite{ngo2024delta}+ZoeDepth & 79.2 & 66.3 \\
MASt3R~\cite{leroy2024grounding} & -- & 40.5 \\
MASt3R~\cite{leroy2024grounding}+ZoeDepth & 36.1 & 33.8 \\
PointSt3R & -- & 61.2 \\
PointSt3R+ZoeDepth & 75.3 & 62.9 \\
\bottomrule
\end{tabular}}
\vspace*{-6pt}
\caption{APD (average points under distance) metric with the fixed thresholds of $[0.1,0.3,0.5,1.0]$ used in \cite{feng2025st4rtrack}, and median global scaling on the PStudio minival set. ``w G.T'': Results with the ground truth intrinsics matrix for lifting to 3D. ``w/o G.T.'': Results without the ground truth intrinsics.}
\label{tab:point_tracking_3d}
\vspace{-12pt}
\end{table}

\subsection{3D Point Tracking} We report results of 3D point tracking on the APD metric for the mini validation dataset of PStudio (from the TAP-Vid-3D~\cite{koppula2024tapvid} benchmark) in Table \ref{tab:point_tracking_3d}. We choose to follow St4RTrack~\cite{feng2025st4rtrack} and use global median scaling, due to scale ambiguity, and static thresholds of $[0.1, 0.3, 0.5, 1.0]$\footnote{We do not compare to St4rTrack itself~\cite{feng2025st4rtrack} as the model is not public.}. All point trackers integrate ZoeDepth~\cite{bhat2023zoedepth} for lifting to 3D. For direct comparison, we also report 3D tracking accuracy for MASt3R and PointSt3R using ZoeDepth. Furthermore, it is standard for the ground truth intrinsics matrix to be used when lifting 2D point trackers to 3D~\cite{koppula2024tapvid}. However, MASt3R and PointSt3R do not have access to these, so we also include results without ground truth intrinsics (see supplementary for extra details).  %

PointSt3R shows a consistent improvement over MASt3R, whether using ZoeDepth ($+39.2\%$ and $+29.1\%$), or pointmaps ($20.7\%$). However, PointSt3R does fall slightly short of the point trackers. With ZoeDepth, PointSt3R underperforms CoTracker3~\cite{karaev2024cotracker3} by $5.0\%$ and $3.3\%$ when using the ground truth intrinsics and when using estimated intrinsics, respectively. 
Importantly, we show that PointSt3R performs comparably with estimated intrinsics with and without depth estimation (61.2 vs 62.9\%).

\subsection{Ablations}
\label{sec:ablations}

\begin{figure*}[!th]
    \centering
   \includegraphics[width=1.0\linewidth]{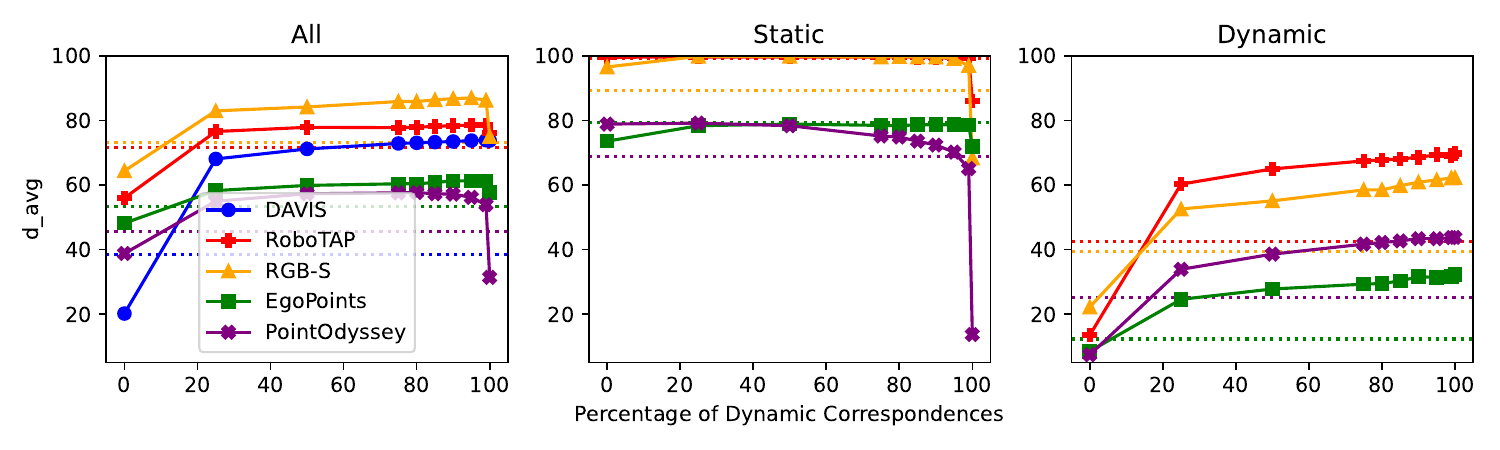}
   \vspace*{-2em}
\caption{Tracking accuracy $\delta_{\text{avg}}\uparrow$ of PointSt3R when changing the percentage of dynamic correspondences $r$, where $0\%$ is only static correspondences and $100\%$ is dynamic correspondences only. The dashed lines represent MASt3R~\cite{leroy2024grounding}.}
    \vspace*{-12pt}
  \label{fig:ablation_ratio_points}
\end{figure*}

\noindent \textbf{Percentage of Dynamic Correspondences.}\label{sec:ablation_all_v_dyn}
Figure \ref{fig:ablation_ratio_points} compares the performance when we change the percentage of dynamic correspondences $r$. This is evaluated on all four 2D tracking datasets of table \ref{tab:point_tracking_2d}, as well as the small split of PointOdyssey~\cite{zheng2023pointodyssey}, described in section \ref{sec:datasets}. The dashed lines show the performance of the MASt3R baseline. It is clear that the model benefits from seeing more dynamic correspondences, as performance on nearly all datasets and splits increases as the percentage of dynamic correspondences increases. 
When training with 100\% dynamic points, catastrophic forgetting of the 3D grounding can be seen with a clear degradation, especially for PointOdyssey, RoboTAP and RGB-S. Interestingly, we see a similar performance drop when using only static correspondences, often dropping below the baseline. We choose $r = 95\%$ as our final PointSt3R model for all other results.

\begin{table}[t]
\centering
\small
\begin{tabular}{ll}
\toprule
Strides  & $\delta_{\text{avg}}\uparrow$ \\
\midrule
$[1,2,3,4,5,6,7,8,9]$ & 68.5  \\
$[1,5,10,15,20,25,30,35,40]$ & 72.2 \\
$[10,20,30,40,50,60,70,80,90]$ & 73.4 \\
$[10,30,50,70,90,110,130,150,170]$ & \textbf{73.8} \\
\bottomrule
\end{tabular}
\caption{Tracking accuracy on TAP-Vid-DAVIS~\cite{doersch2022tap} when changing temporal strides on PointSt3R training.}
\vspace*{-12pt}
\label{tab:ablation_strides}
\end{table}

\begin{figure}[t]
  \centering
   \includegraphics[width=0.9\linewidth]{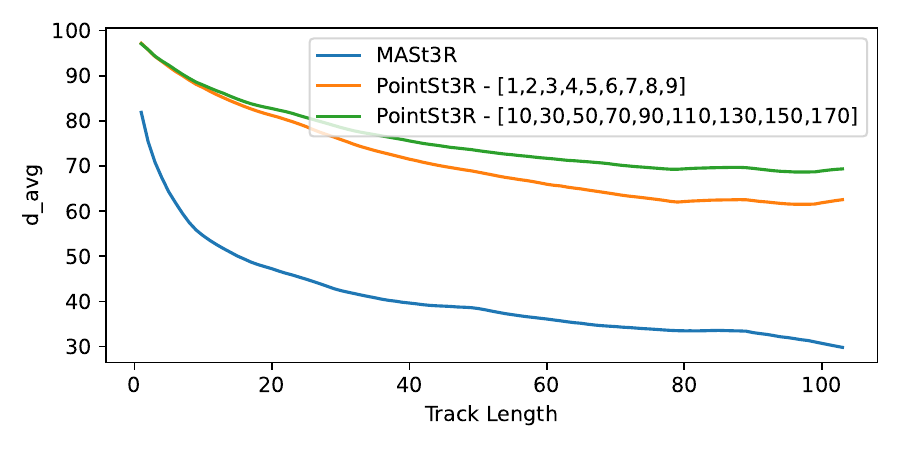}
   \vspace*{-1em}
   \caption{Comparison of tracking accuracy $\delta_{\text{avg}}\uparrow$ on TAP-Vid-DAVIS~\cite{doersch2022tap} against track length (in frames) for MASt3R~\cite{leroy2024grounding} and PointSt3R (Dynamic data) with two stride options for training. Green line shows the advantage of bigger strides for longer tracks.}
   \vspace*{-6pt}
   \label{fig:ablation_stride_accuracy}
\end{figure}

\vspace{1.0em}\noindent \textbf{Impact of Stride Length.} We conduct two ablations clarifying the impact of the stride lengths of training data pairs.
First, we show the tracking accuracy for different stride options during training, in Table~\ref{tab:ablation_strides}. 
The results clearly show that larger strides produce much better results on dynamic tracking. 

We also compare PointSt3R with another version trained with the smallest stride option of Table \ref{tab:ablation_strides} for all three datasets and report the cumulative tracking accuracy on TAP-Vid-DAVIS over the different track lengths. It is expected that longer tracks are harder for point tracking in general and correspondence specifically. Figure \ref{fig:ablation_stride_accuracy} shows that although tracking accuracy is roughly similar for the two variants of PointSt3R for short tracks, there is a clear advantage to the larger stride training for tracks longer than 20 frames. This confirms that larger strides benefit longer-term point tracking.

\begin{table}[t]
\centering
\small
\begin{tabular}{ll}
\toprule
Datasets  & $\delta_{\text{avg}}\uparrow$ \\
\midrule
PO~\cite{zheng2023pointodyssey} & 72.7  \\
PO~\cite{zheng2023pointodyssey}+Kubric~\cite{karaev2024cotracker3} & 73.2 \\
PO~\cite{zheng2023pointodyssey}+Kubric~\cite{karaev2024cotracker3}+DynamicReplica~\cite{karaev2023dynamicstereo} & 73.8 \\
\bottomrule
\end{tabular}
\vspace*{-6pt}
\caption{Tracking accuracy on TAP-Vid-DAVIS~\cite{doersch2022tap} when changing datasets on fine-tuning.}
\label{tab:ablation_datasets}
\vspace*{-16pt}
\end{table}

\vspace{1.0em}\noindent \textbf{Impact of Training Datasets.} Table \ref{tab:ablation_datasets}  reports the contribution of all three training datasets, measured in terms of tracking accuracy on TAP-Vid-DAVIS~\cite{doersch2022tap}. Results compare training on PointOdyssey alone, training on PointOdyssey plus Kubric, and training on PointOdyssey plus Kubric plus DynamicReplica. Note that we only train with $10,000$ samples for each model, but this is equally distributed across the training datasets (i.e. 10k for PO, 5k each for PO+Kubric and roughly 3.3k each for PO+Kubric+DynamicReplica). Including a diverse range of datasets improves tracking accuracy. We use all three for all other results.

\vspace{-4pt}
\section{Conclusion}
\vspace{-6pt}
In this paper, we introduce PointSt3R, a dynamic correspondence-focused variant of MASt3R~\cite{leroy2024grounding} that achieves competitive performance in both 2D and 3D point tracking when compared to recent models. We note that the large-scale training of MASt3R helps it correspond fairly well in the static scene, yet it fails to track dynamic objects. Inspired by MonSt3R~\cite{zhang2024monst3r}, which showed that (correspondence-free) dynamic scene reconstruction can be substantially improved with targeted training on dynamic scenes, PointSt3R explores how general-purpose correspondence can be improved through training as well. 
We introduce a new dynamic matching loss and a visibility head, and show that PointSt3R competes well with state-of-the-art point trackers on multiple datasets, even though PointSt3R does not use temporal context. Our results demonstrate that large-scale 3D grounded approaches to correspondence, even operating 2 frames at a time, are useful not only for 3D reconstruction but for tracking in complex, dynamic scenes.

\small{\noindent \textbf{Acknowledgments:} This work is supported by EPSRC Doctoral Training Program, EPSRC UMPIRE EP/T004991/1 and EPSRC Programme Grant VisualAI EP/T028572/1. We acknowledge the use of the Isambard-AI National AI Research Resource (AIRR) funded by DSIT [ST/AIRR/I-A-I/1023]. Meta only served in an advisory capacity and no experiments were conducted at/by Meta.}

{\small
\bibliographystyle{ieee_fullname}
\bibliography{egbib}
}

\clearpage
\newpage
\renewcommand{\thesubsection}{\Alph{subsection}}
\section*{Appendix}
\subsection{Visibility}
\begin{figure}[t]
  \centering
   \includegraphics[width=\linewidth]{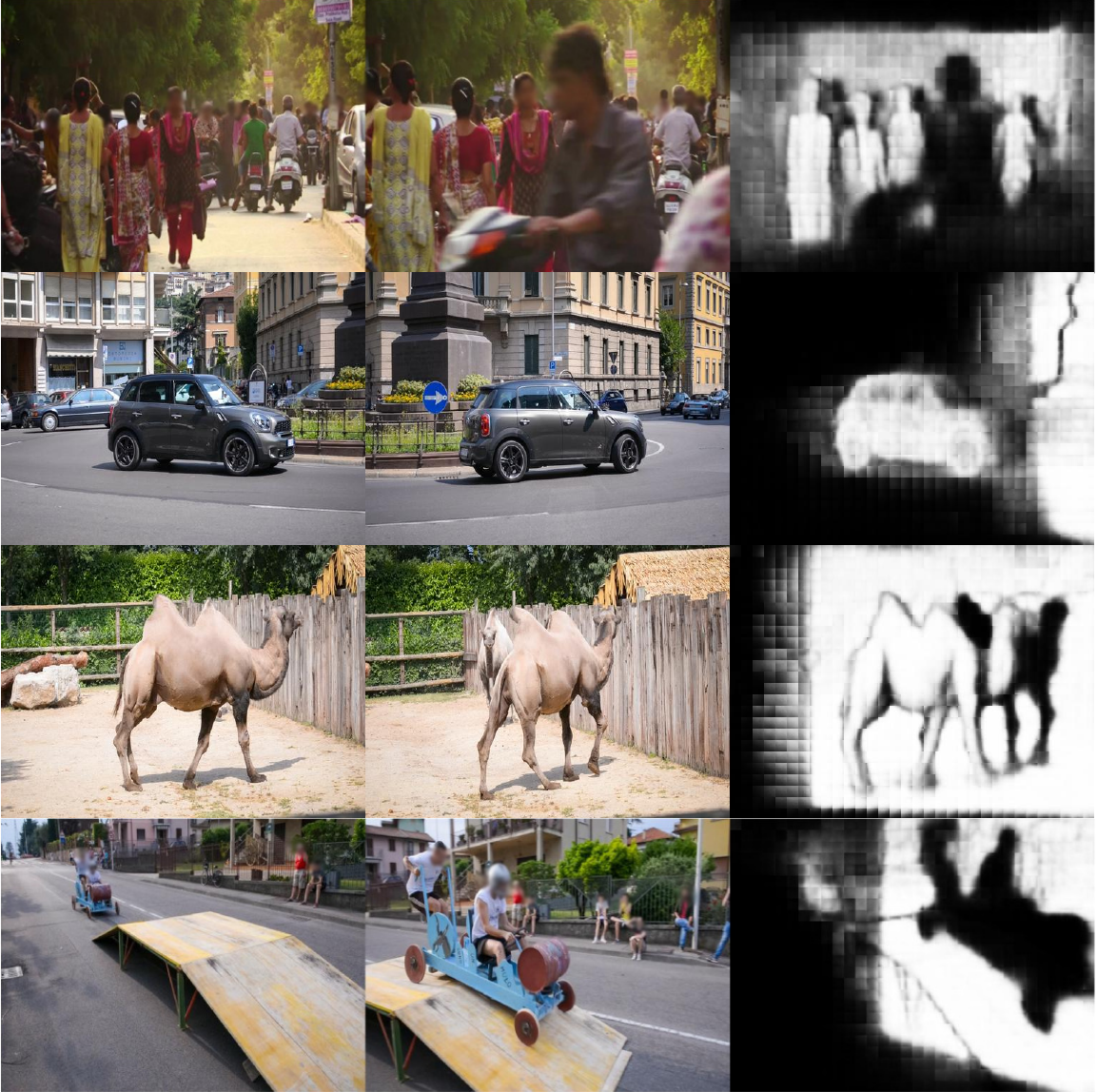}
   \vspace{-1.5em}
   \caption{Visualisation of the visibility heatmaps produced by our added visibility prediction head in the query branch. Each head predicts whether each pixel is visible or not in the other branch's image. Here white is high probability and black is low.}
   \vspace{-1em}
   \label{fig:vis_vis}
\end{figure}

In figure \ref{fig:vis_vis}, we present qualitative examples of the visibility heatmaps produced by the query image branch as these are used to predict visibility for our model. As can be seen, the visibility prediction heads do a good job at identifying which points are visible and occluded in the target image.
Specifically, in the first row, the rider in the foreground~(middle) occludes the people seen in the query image~(left) and this is correctly predicted by the black region in the centre of the frame~(right). Similarly, in the last row, the points on the cart~(left) are predicted correctly as visible~(right).
Conversely, the part of the scene that disappears due to the camera pan to the right is clearly classified as occluded with a low probability.
It should also be noted that the visibility head correctly captures the visibility due to the camera motion. 
In the first, third and fourth rows, the part of the scene visible in the second image is easily distinguished in the visibility map.
In the second row, only the column is common between frames and is predicted as visible.

\subsection{Resolution and Bilinear Interpolation}
In training, MASt3R and PointSt3R have multiple resolution options, with the x-axis kept constant at 512 pixels. We conduct an ablation of the affect of changing input resolution at inference on 2D tracking for TAP-Vid-DAVIS. As expected, the largest input resolution performs best and this happens to be a common native resolution for point trackers. We also report how results change between using the nearest pixel's feature to the query point and when using bilinear interpolation to query the correspondence feature. As expected bilinear interpolation produces more accurate results, with an increase of $1.1\%$.

\begin{table}[h!]
\centering
\small
\begin{tabular}{lll}
\toprule
Resolution & Sampling & $\delta_{\text{avg}}\uparrow$ \\
\midrule
$(512, 160)$ & Nearest  & 63.5   \\
$(512, 256)$ & Nearest  & 70.0   \\
$(512, 288)$ & Nearest  & 70.7 \\
$(512, 336)$ & Nearest  & 71.9 \\
$(512, 384)$ & Nearest  & \underline{72.7} \\
$(512, 384)$ & Bilinear & \textbf{73.8} \\
\bottomrule
\end{tabular}
\caption{Tracking accuracy $\delta_{\text{avg}}\uparrow$ on TAP-Vid-DAVIS~\cite{doersch2022tap} when changing resolution and sampling strategy. "Nearest" is simply selecting the nearest pixel's feature to the query coordinate. "Bilinear" uses bilinear interpolation to retrieve the feature for matching.}
\vspace*{-12pt}
\label{tab:ablation_computation}
\end{table}

\begin{table*}[b]
\centering
\resizebox{0.92\linewidth}{!}{
\begin{tabular}{l|ccccc|cccc|cccc}
\toprule
           &                \multicolumn{5}{c|}{All Points} & \multicolumn{4}{c|}{Static Points} & \multicolumn{4}{c}{Dynamic Points} \\
\cmidrule{2-14}
Model      & TAP-Vid-DAVIS & RoboTAP  & RGB-S & EgoPoints & PO & RoboTAP & RGB-S & EgoPoints & PO &  RoboTAP & RGB-S & EgoPoints & PO  \\
\midrule
{MASt3R}~\cite{leroy2024grounding} & 38.5 & 71.6 & 73.2 & 53.5 & 45.6 & 99.2 & 89.3 & 79.4 & 68.8 & 42.5 & 39.4 & 12.3 & 25.2 \\
\midrule
PointSt3R - $0\%$                  & 20.2 & 56.0 & 64.4 & 48.2 & 38.8 & 99.6 & 96.6 & 73.6 & 78.9 & 13.5 & 22.2 &  8.5 &  7.4 \\
PointSt3R - $25\%$                 & 68.1 & 76.6 & 83.0 & 58.3 & 55.0 & 99.7 & 99.9 & 78.4 & 79.2 & 60.3 & 52.6 & 24.6 & 33.9 \\
PointSt3R - $50\%$                 & 71.2 & 77.9 & 84.2 & 59.9 & 57.3 & 99.7 & 99.9 & 78.9 & 78.4 & 65.0 & 55.1 & 27.8 & 38.6 \\
PointSt3R - $75\%$                 & 72.9 & 77.8 & 85.9 & 60.4 & 57.6 & 99.7 & 99.7 & 78.5 & 75.2 & 67.5 & 58.5 & 29.3 & 41.7 \\
PointSt3R - $80\%$                 & 73.1 & 77.9 & 85.9 & 60.4 & 57.7 & 99.8 & 99.8 & 78.4 & 74.9 & 67.8 & 58.5 & 29.5 & 42.2 \\
PointSt3R - $85\%$                 & 73.3 & 78.2 & 86.4 & 60.8 & 57.3 & 99.6 & 99.7 & 78.7 & 73.6 & 68.1 & 59.8 & 30.3 & 42.7 \\
PointSt3R - $90\%$                 & 73.5 & 78.4 & 86.8 & 61.3 & 57.2 & 99.5 & 99.7 & 78.7 & 72.4 & 68.6 & 60.9 & 31.6 & 43.4 \\
PointSt3R - $95\%$                 & 73.8 & 78.6 & 87.0 & 61.3 & 56.2 & 99.4 & 99.2 & 78.7 & 70.2 & 69.4 & 61.6 & 31.4 & 43.4 \\
PointSt3R - $99\%$                 & 73.9 & 78.6 & 86.3 & 61.3 & 53.9 & 99.5 & 97.0 & 78.6 & 65.1 & 69.2 & 62.2 & 31.7 & 43.7 \\
PointSt3R - $100\%$                & 73.8 & 76.2 & 75.0 & 57.7 & 31.4 & 86.1 & 68.4 & 71.9 & 13.7 & 69.8 & 62.3 & 32.3 & 43.8 \\
\bottomrule
\end{tabular}
}
\caption{Tracking accuracy $\delta_{\text{avg}}\uparrow$ for PointSt3R models with different percentage of dynamic correspondences per batch, compared with the MASt3R baseline.}
\label{tab:ratio_ablation_table}
\vspace{-1.0em}
\end{table*}

\subsection{Percentage of Dynamic Correspondences}
We include the full table of results in table \ref{tab:ratio_ablation_table} for the figure of the ablation of percentage of dynamic correspondences in each batch. It should be noted that our chosen model was $95\%$ for its consistent performance across datasets and splits.

\begin{table}[h!]
\centering
\small
\begin{tabular}{ll}
\toprule
Model  & Time (s) \\
\midrule
CoTracker3~\cite{karaev2024cotracker3} & 0.3  \\
PointSt3R                              & 1.7 \\
\bottomrule
\end{tabular}
\caption{Inference time for CoTracker3~\cite{karaev2024cotracker3} and PointSt3R for a window for 16 frames and 5 query points.}
\vspace*{-12pt}
\label{tab:ablation_computation}
\end{table}

\subsection{Computational Cost}
PointSt3R uses global cosine similarity and produces results in a pairwise fashion. It is not particularly optimised. 
We compare the runtime to CoTracker3~\cite{karaev2024cotracker3} which uses a window of 16 frames, so we chose to feed PointSt3R a batch of 16 image pairs and performed cosine similarity to produce tracks for 5 query points and then compared the runtime. As can be seen from table \ref{tab:ablation_computation}, PointSt3R is 5x slower than CoTracker3, which is expected due to the pairwise method used.

We highlight that our attempt is not to produce a competitive point tracker in performance, but to evaluate the capabilities of current 3D grounding foundation models for the task of point tracking.
Our results also highlight that the temporal context can be removed from point trackers without impacting the performance much.

\subsection{3D Tracking - Implementation Details}
For the 3D tracking results seen in the main paper, we employ ZoeDepth~\cite{bhat2023zoedepth} NK model. When the ground truth intrinsics are not used, we use the demo.py, available with the SpaTracker~\cite{xiao2024spatialtracker} published code, that estimates the intrinsics for videos without ground truth by setting $f_x = f_y = W$ and $c_x = W/2, c_y = H/2$.

\end{document}